\documentclass[reprint, amsmath, amssymb, aps,groupedaddress,superscriptaddress]{revtex4-2}

\usepackage{amsmath}
\usepackage{amsfonts}
\usepackage{amssymb}
\usepackage{graphicx}
\usepackage{color}
\usepackage{ragged2e}
\usepackage{comment}
\usepackage{hyperref}
\usepackage{cleveref}
\usepackage{float}
\usepackage{caption}

\begin{document}

\title{Achieving Occam's Razor: Deep Learning for Optimal Model Reduction}

\author{Botond B. Antal}
\affiliation{Department of Biomedical Engineering, State University of New York at Stony Brook, Stony Brook, NY, USA}
\author{Anthony G. Chesebro}
\affiliation{Department of Biomedical Engineering, State University of New York at Stony Brook, Stony Brook, NY, USA}
\author{Helmut H. Strey}
\affiliation{Department of Biomedical Engineering, State University of New York at Stony Brook, Stony Brook, NY, USA}
\affiliation{Laufer Center for Physical and Quantitative Biology, State University of New York at Stony Brook, Stony Brook, NY, USA}
\author{Lilianne R. Mujica-Parodi}
\affiliation{Department of Biomedical Engineering, State University of New York at Stony Brook, Stony Brook, NY, USA}
\affiliation{Laufer Center for Physical and Quantitative Biology, Stony Brook University, Stony Brook, NY, USA}
\author{Corey Weistuch}
\email[Author for correspondence: ]{weistucc@mskcc.org}
\affiliation{Department of Medical Physics, Memorial Sloan Kettering Cancer Center, New York, NY, USA}

\vspace{10pt}

\date{\today}

\begin{abstract}
All fields of science depend on mathematical models. \textit{Occam's razor} refers to the principle that good models should exclude parameters beyond those minimally required to describe the systems they represent. This is because redundancy can lead to incorrect estimates of model parameters from data, and thus inaccurate or ambiguous conclusions. Here, we show how deep learning can be powerfully leveraged to address Occam's razor. FixFit uses a feedforward deep neural network with a bottleneck layer to characterize and predict the behavior of a given model from its input parameters. FixFit has three major benefits.  First, it provides a metric to quantify the original model's degree of complexity.  Second, it allows for the unique fitting of data.  Third, it provides an unbiased way to discriminate between experimental hypotheses that add value versus those that do not. In two use cases, we demonstrate the broad applicability of this method across scientific domains. To validate the method using a known system, we apply FixFit to recover known composite parameters for the Kepler orbit model. To illustrate how the method can be applied to less well-established fields, we use it to identify parameters for a multi-scale brain model and reduce the search space for viable candidate mechanisms.
\end{abstract}

\maketitle


\justifying


\section{Introduction}

Mathematical models are commonly used to describe the dynamical behavior of physical systems. Yet model parameters are not mere mathematical descriptions.  The accurate estimation of parameter values can often yield deep mechanistic insight, whether with respect to the properties of particles \cite{gfitter2012updated}, interactions among genetic networks \cite{d1999linear}, or the generation of neuronal signaling \cite{hodgkin1952currents}. 

A fundamental challenge in parameter fitting and the construction of models stems from parameter redundancies\cite{prinz2004similar, kacprzak2022deeplss}. Parameter degeneracy is a particularly problematic in multi-scale models, where emergent measured values exist many layers above their mechanistic parameters. This can lead to many different combinations of parameters fitting the observed data equally well, a phenomenon described as \textit{overdetermined} or ``sloppy models'' \cite{gutenkunst2007universally, chis2011structural, transtrum2015perspective}. Finding these non-unique solutions is also difficult, as parameter interactions can introduce numerous local minima that hinder algorithms for data fitting \cite{weise2009global}. Consequently, there is often a trade-off between using an easier-to-interpret model with fewer parameters that fails to describe the system accurately and using a highly-detailed model that risks redundancy of its parameters \cite{ryom2023speed}.

A solution to this trade-off is to identify and account for parameter dependencies. One of the most widely-used tools to quantify interactions among parameters is the Fisher Information Matrix, which can be acquired through a nonlinear least-square Levenberg-Marquardt algorithm. This procedure, for given data, finds the locally best-fitting parameter values and their covariance matrix \cite{more2006levenberg, transtrum2015perspective}. In overdetermined models, these covariances are strong for pairs of parameters that are not separable. While the Fisher Information Matrix only characterizes linear interactions, other methods can also uncover nonlinear interactions \cite{chis2011structural, wieland2021structural}. For example, one can determine parameter interactions by approximating the model's behavior as a function of its parameters using the Taylor series expansion \cite{pohjanpalo1978system}. Nevertheless, a common limitation of these methods is that they do not provide a functional form for the redundancies, and therefore cannot guide the parameter fitting process. A sensible strategy is to combine the redundant parameters into composite parameters with a unique best fit. Although suitable methods exist, they rely on studying the model analytically or through its local behaviors and derivatives \cite{cole2020parameter}, a strategy that would not be feasible for highly complex models. Thus, there remains a need for a more general approach.

\begin{figure*}[!t]
    \begin{center}
	    \includegraphics[scale=0.65]
     {
	    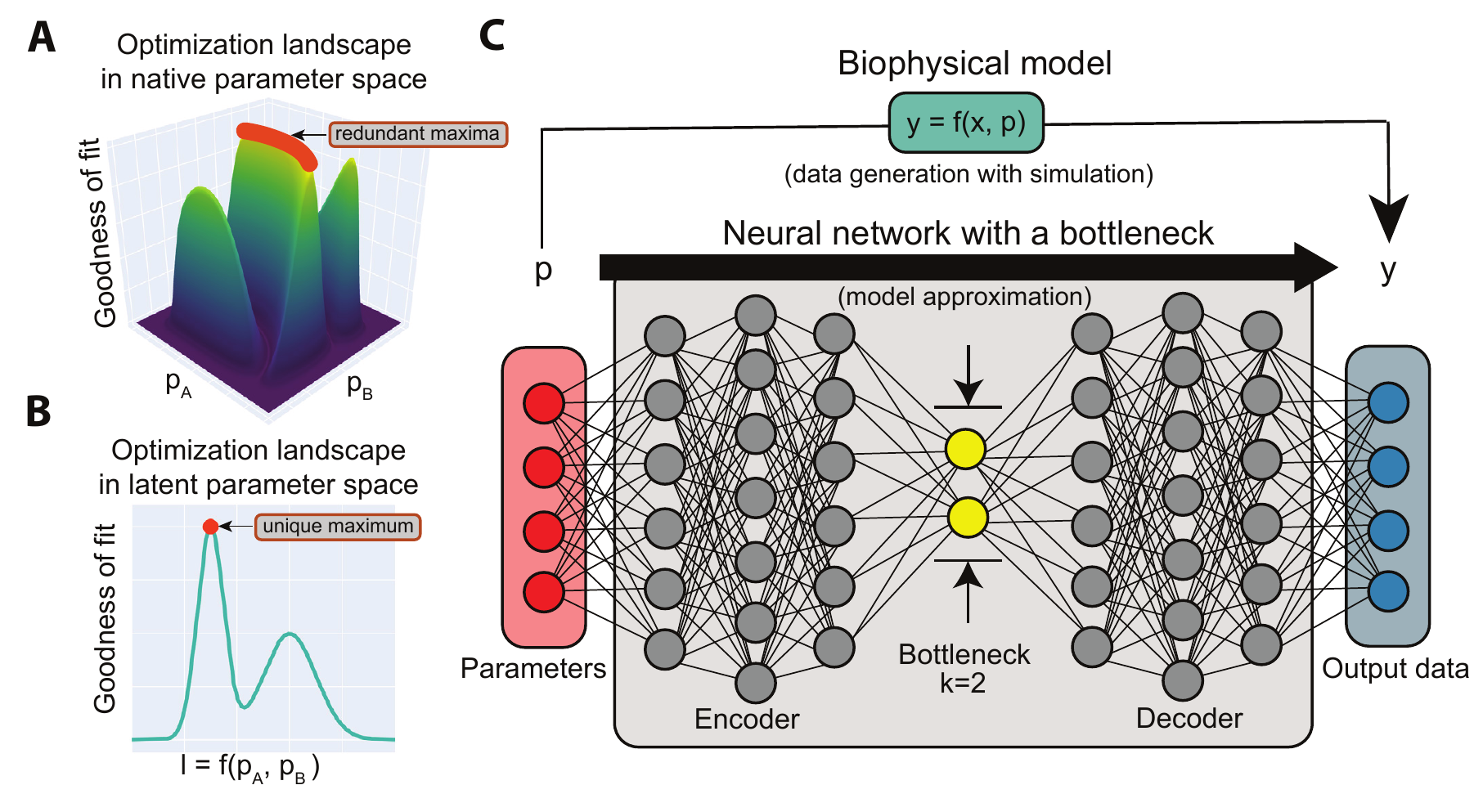}
    \end{center}
	\caption{\small\raggedright{{{\textbf{FixFit compresses interacting parameters into a latent representation that can be uniquely inferred from data.} \textbf{A}: A schematic representing the goodness of fit landscape of a model with two interacting parameters. These interactions cause multiple parameter combinations to fit experimental data equally (the redundant maxima in red). \textbf{B}: The same landscape but with the two interacting parameters first combined into a single latent variable. In contrast to the native parameters, numerical fitting over latent variables will converge to a unique solution. \textbf{C}: FixFit generates such unique latent representations using a neural network with an encoder, bottleneck layer, and a decoder. After determining the optimal number ($k$) of latent (bottleneck) nodes (see Methods), the neural network is trained on pairs of parameters and their corresponding outputs from computational simulations of a model of interest. Following training, the bottleneck layer will include a representation of input parameters that is uniquely inferable from output data. With the latent representation established, the decoder section of the neural network can be combined with an optimizer to infer unique parameters in latent space from previously unseen output data. In addition, the encoder part can be combined with sensitivity analysis to determine the influence of input parameters on the latent representation. This enables us to characterize changes in different output samples in terms of underlying parameters that would not otherwise be accessible.}}}}
 
	\label{fig:scheme}
\end{figure*}

Here we show how one may identify and estimate the largest set of lower-dimensional latent parameters uniquely resolved by the data (\autoref{fig:scheme}). The framework, \textit{FixFit}, builds on previously established methods and consists of three steps. First, we find these latent parameters using a neural network with a bottleneck \cite{kramer1991nonlinear}. We optimize the same architecture at variable bottleneck widths to identify the optimal latent dimension \cite{tishby2000information, tishby2015deep, achille2017emergence}. The ability of neural networks to approximate any function allows our method to be applied to models of arbitrary complexity \cite{hornik1989multilayer, hornik1991approximation, csaji2001approximation}. Next, FixFit establishes the relationship between latent and original parameters using global sensitivity analysis \cite{li2010global}. Finally, we fit latent parameters to data using global optimization techniques \cite{virtanen2020scipy}.

Here we provide two use-cases for FixFit.  To establish validity against a known system, we first demonstrate its ability to recover nonlinear parameter combinations for the Kepler orbit model \cite{bate2020fundamentals}. To demonstrate its potential for scientific discovery, we then use FixFit to identify previously undiscovered parameter redundancies in the multi-scale Larter-Breakspear neural mass model \cite{larter1999coupled, breakspear2003modulation}.

\section{Methods}

\subsection{Models}
\subsubsection{Kepler orbit model}

The classical Kepler orbit model captures the gravitational motion of two orbiting planetary bodies through four input parameters (\autoref{fig:kepler}A) \cite{bate2020fundamentals}: the mass of the orbiting body ($m_1$), the mass of the body fixed at the focus point ($m_2$), the closest distance between the two bodies ($r_0$) and the corresponding initial angular velocity ($\omega_0$). To facilitate later training of a neural network, the units of each parameter were chosen such that the parameter values had comparable magnitudes. Thus, $m_1$ and $m_2$ were expressed in $kg$, $r_0$ in $m$, and $w_0$ in $day^{-1}$. The center of mass of the orbiting body evolves according to:

\begin{equation} \label{eq:kepler_origin}
    m_1 \frac{d^2r}{dt^2} - m_1 r \omega^2=\frac{G m_1 m_2}{r^2}
\end{equation}

where $G$ is the universal constant of gravitation (approximately $0.5 m^3 kg^{-1} day^{-2}$), $r$ is the distance between the two bodies, and $\omega$ is the angular velocity $\frac{d\theta}{dt}$. As is apparent, the mass of the orbiting body ($m_1$) cancels in the above equation and thus cannot be resolved from data. More broadly, the dynamics are highly constrained by the conservation of energy and angular momentum. As a result, the orbit $r$ can be readily solved as a function of the angle $\theta \in [0,2\pi)$ swept by the orbiting object (starting at $\theta=0$) \cite{bate2020fundamentals}. These solved orbits are simple ellipses and thus can be described using only two pieces of information, the eccentricity $e$ (the shape) and the semi-latus rectum $l$ (the size), as follows:

\begin{equation} \label{eq:kepler_orbit}
    r(\theta)=\frac{l}{1+e \cos(\theta)}
\end{equation}

with

\begin{equation} \label{eq:kepler_e_short}
    e = \bigg |\frac{r_0^3 \omega_0^2}{G m_2} - 1 \bigg |
\end{equation}

\begin{equation} \label{eq:kepler_l_short}
    l = \frac{r_0^4 \omega_0^2}{G m_2}
\end{equation}

The above two equations establish a ground truth for the degeneracy that we aimed to characterize and overcome during parameter inference. Specifically, parameter $m_1$ is canceled out and is therefore a completely redundant parameter. The remaining three parameters ($m_2$, $r_0$, $\omega_0$) can be compressed into two terms and still uniquely map to outputs \cite{bate2020fundamentals}.

\subsubsection{Multi-scale brain model}
The Larter-Breakspear model (\autoref{fig:lbscheme}) describes the dynamics of a group of coupled brain regions, each modeled as a population of neurons \cite{larter1999coupled, breakspear2003modulation, endo2020evaluation, chesebro2023ion}. Each brain region captures the averaged synaptic processes and voltage-dependent ion transport ($\text{K}^+$, $\text{Na}^+$, and $\text{Ca}^{2+}$) of its constituent neurons. These effective dynamics are described through three state variables: mean excitatory membrane voltage V(t), mean inhibitory membrane voltage Z(t), and the proportion of open potassium channels W(t). Note that although $V$, $Z$, and $W$ depend on time, this notation is omitted in the following equations for clarity. These states of each $i$th brain region evolve according to:

\begin{multline}
\frac{dV_i}{dt} = \\ -(g_\text{Ca} + r_{\text{NMDA}} a_{ee} [(1 - c) Q_V + c Q_i^{\text{network}}]) m_\text{Ca} (V_i - V_\text{Ca}) \\
        -(g_\text{Na} m_\text{Na} + a_{ee}  [(1 - c) Q_V + c Q_i^{\text{network}}]) (V_i - V_\text{Na}) \\
        -g_\text{K} W_i (V_i - V_\text{K}) - g_L (V_i - V_L) - a_{ie} Z_i Q_z + a_{ne} I_0
\end{multline}

\begin{equation}
    \frac{dZ_i}{dt} = b (a_{ni} I_0 + a_{ei} V_i Q_V)
\end{equation}

\begin{equation}
    \frac{dW_i}{dt} = \phi \frac{m_\text{K} - W_i}{\tau_\text{K}}
\end{equation}

Where $Q_V$/$Q_Z$ are excitatory/inhibitory mean firing rates, and $m_\text{Na}$, $m_\text{K}$, $m_\text{Ca}$ are ion channel gating functions. These are computed as:

\begin{equation}
    Q_V = 0.5 Q_{V_{\text{max}}} (1 + \tanh(\frac{V - V_T}{\delta}))
\end{equation}

\begin{equation}
    Q_Z = 0.5 Q_{Z_{\text{max}}} (1 + \tanh(\frac{V - V_Z}{\delta}))
\end{equation}

\begin{equation}
    m_{\text{ion}} = 0.5 (1 + \tanh(\frac{V - T_{\text{ion}}}{\delta_{\text{ion}}}))
\end{equation}

Brain regions are connected through the coupling term $Q_i^{\text{network}}$, which is scaled with a global coupling constant $c$, to produce whole-brain-scale dynamics. $Q_i^{\text{network}}$ is given by:

\begin{equation}
    Q_i^{\text{network}} = \frac{\sum_j u_{i,j} Q_{V_j}}{\sum u_{i,j}}
\end{equation}

We considered 78 brain regions selected from the Desikan-Killiany atlas included in FreeSurfer, with inter-regional structural connectivity ($u_{i,j}$) that was determined from diffusion tensor imaging (DTI) data from 13 healthy human adults \cite{desikan2006automated, endo2020evaluation}. All parameters of the Larter-Breakspear model are described in \autoref{tab:lb_pars}.

\begin{table*}[!t]
\caption{Parameters of the Larter-Breakspear multi-scale brain model\label{tab:lb_pars}}
\centering
\begin{tabular}{ |p{3cm}|p{10cm}|  }

\hline
\textbf{Parameter} & \textbf{Description} \\
\hline
$V_\text{Na}$ & $\text{Na}^+$ reversal potential \\
$V_\text{K}$ & $\text{K}^+$ reversal potential \\
$V_\text{Ca}$ & $\text{Ca}^{2+}$ reversal potential \\
$V_\text{L}$ & Leak channels reversal potential \\
$g_\text{Na}$ & $\text{Na}^+$ conductance \\
$g_\text{K}$ & $\text{K}^+$ conductance \\
$g_\text{Ca}$ & $\text{Ca}^{2+}$ conductance \\
$g_\text{L} $ & Leak channels conductance \\
$T_\text{Na}$ & $\text{Na}^+$ channel threshold \\
$T_\text{K}$ & $\text{K}^+$ channel threshold \\
$T_\text{Ca}$ & $\text{Ca}^{2+}$ channel threshold \\
$\delta_\text{Na}$ & $\text{Na}^+$ channel threshold variance \\
$\delta_\text{K}$ & $\text{K}^+$ channel threshold variance \\
$\delta_\text{Ca}$ & $\text{Ca}^{2+}$ channel threshold variance \\
$V_T$ & Excitatory neuron threshold voltage \\
$Z_T$ & Inhibitory neuron threshold voltage \\
$\delta$ & Variance of thresholds \\
$Q_{V_\text{max}}$ & Excitatory population maximum firing rate \\
$Q_{Z_\text{max}}$ & Inhibitory population maximum firing rate \\
$a_{ee}$ & Excitatory-to-excitatory synaptic strength \\
$a_{ei}$ & Excitatory-to-inhibitory synaptic strength \\
$a_{ie}$ & Inhibitory-to-excitatory synaptic strength \\
$a_{ne}$ & Non-specific-to-excitatory synaptic strength \\
$a_{ni}$ & Non-specific-to-inhibitory synaptic strength \\
$I_0$ & Subcortical excitatory input \\
$b$ & Time scaling factor \\
$\phi$ & Temperature scaling factor \\
$\tau_\text{K}$ & $\text{K}^+$ relaxation time \\
$r_\text{NMDA}$ & NMDA/AMPA receptor ratio \\
$c$ & Global region-to-region coupling constant \\
\hline
\end{tabular}
\end{table*}

Next, to produce a signal compatible with functional MRI, a commonly utilized neuroimaging modality, the simulated region-specific excitatory signals were transformed into blood-oxygen-level-dependent (BOLD) signal via the Balloon-Windkessel model with standard parameters \cite{endo2020evaluation, friston2003dynamic}. Finally, we derived functional connectivity (FC) from a simulated BOLD signal to yield a phase-invariant signal. We quantified FC with all-to-all Pearson correlation coefficients among the 78 regions, resulting in a 78-by-78 correlation matrix for each simulation.

\subsection{Data generation}
\subsubsection{Kepler orbit model}
Training and validation data for the neural network were generated using equations \ref{eq:kepler_orbit} - \ref{eq:kepler_l_short}. The four input parameters, $m_1$, $m_2$, $r_0$, and $\omega_0$, were sampled with a four dimensional Sobol sequence (SciPy v1.7.1 \cite{virtanen2020scipy}). All four parameters were drawn from a range of $[0.1, 1]$ (Figure S1 \cite{SI}). Next, a subset of samples was rejected based on an eccentricity criterion. A raw parameter set was discarded if $e > 0.95$ or $e < 0.7$. These two conditions ensured all resultant orbits were ellipse-shaped with moderate eccentricity. The final sample sizes were 2,276 for training and 253 for validation. Output space consisted of $r(\theta)$ values computed at 100 evenly distributed $\theta$ values within the range of $[0, 2\pi]$ for each input parameter set. The computed $r(\theta)$ values were log-transformed to narrow down their range, thus ensuring a favorable output space for the neural network (Figure S2 \cite{SI}).

\subsubsection{Multi-scale brain model}
We first applied domain knowledge to reduce the 30 parameters of the Larter-Breakspear model to a smaller subset of parameters that is more tractable for parameter inference. We fixed 19 parameters that we determined to be unlikely to be sensitive to different biological conditions to default values. All default values were taken from \cite{endo2020evaluation}. The remaining eleven input parameters of the model were drawn from biologically relevant ranges (\autoref{tab:lb_par_ranges}) using a Sobol sequence. The model was simulated with input parameters still on their original scales. Later, all eleven input parameters were scaled to within a range of $[0, 1]$ for training the neural network (Figure S3 \cite{SI}). Simulations were performed using Neuroblox.jl, a Julia library optimized for high-performance computing of dynamical brain circuit models (http://www.neuroblox.org); while the Larter-Breakspear model is technically unitless, the parameters are scaled so that a single timestep is 1 ms. \cite{breakspear2003modulation}. Simulated time series were converted to BOLD using the Balloon-Windkessel model \cite{friston2003dynamic} $(T_R=0.8\text{s})$ \cite{mujica2020diet} and then bandpass filtered ($0.01 < f < 0.1 \: Hz$) \cite{biswal1996reduction} to quantify functional connectivity \cite{biswal1995functional}. We computed functional connectivity among the 78 brain regions from the processed time series and retained values above the diagonal to discard duplicates. The resultant 3,003 Pearson correlation values per sample constituted the output space for the neural network (Figure S4 \cite{SI}). Simulated data were subject to two exclusion criteria to ensure biologically realistic behavior: time series with non-oscillatory behavior or with a mean FC larger than $0.3$ were discarded. As a result, there were 4,730 retained samples for training and 526 for validation.

\begin{table*}[!t]
\caption{Investigated parameter subset of the Larter-Breakspear multi-scale brain model\label{tab:lb_par_ranges}}
\centering
\begin{tabular}{ |p{3cm}|p{11.5cm}|p{3.cm}|  }
\hline
\textbf{Parameter} & \textbf{Description} & \textbf{Range} \\
\hline
c & Global region-to-region coupling constant & $[0.2, 0.5]$ \\
$\delta$ & Variance of thresholds & $[0.64, 0.7]$ \\
$g_\text{Ca}$ & $\text{Ca}^{2+}$ conductance & $[0.95, 1.05]$ \\
$V_\text{Ca}$ & $\text{Ca}^{2+}$ reversal potential & $[0.95, 1.01]$ \\
$g_\text{K}$ & $\text{K}^+$ conductance & $[1.95, 2.05]$ \\
$V_\text{K}$ & $\text{K}^+$ reversal potential & $[-0.75, -0.65]$ \\
$g_\text{Na}$ & $\text{Na}^+$ conductance & $[6.6, 6.8]$ \\
$V_\text{Na}$ & $\text{Na}^+$ reversal potential & $[0.48, 0.58]$ \\
$a_{ee}$ & Excitatory-to-excitatory synaptic strength & $[0.33, 0.39]$ \\
$a_{ei}$ & Excitatory-to-inhibitory synaptic strength & $[1.95, 2.05]$ \\
$r_\text{NMDA}$ & NMDA/AMPA receptor ratio & $[0.20, 0.30]$ \\
\hline
\end{tabular}
\end{table*}

\subsection{Neural network}
\subsubsection{Kepler orbit model}
We utilized a network with a fully connected architecture and a bottleneck layer in the middle to enforce compression. The network structure was determined empirically based on a trade-off between model accuracy and training speed. There were two hidden layers before and two after the bottleneck layer, all of which had a $\tanh$ activation function. The final output layer, by contrast, was given a linear activation function to map to the output space. By the universal approximation theory of neural networks, hidden layers before and after the bottleneck layer had 14 and 110 nodes \cite{lu2017expressive, kidger2020universal}. The neural network was implemented with TensorFlow (version 2.6.0 \cite{abadi2015tensorflow}). For exact details of the applied architecture, see Figure S5 \cite{SI}.

\subsubsection{Multi-scale brain model}
A similar architecture was used for the brain model with minor adaptations to a significantly wider output space (3,003 values per sample). The encoder included two fully connected hidden layers with 21 nodes in each, whereas the decoder had a single hidden layer with 3,013 nodes. All hidden layers were given $ReLu$ activation, whereas the bottleneck and output layers were both implemented with a linear activation function. Details of the network are described in Figure S6 \cite{SI}.

\subsection{Bottleneck analysis}
To determine the number of uniquely resolvable latent parameters for a given model, we trained our neural network architecture on the previously described training data at varying bottleneck layer dimensions ($k \in \{1,2,3,4,5\}$). At each $k$ increment, ten replicate training runs were performed independently. We employed 5,000 epochs during training to ensure model accuracy was not limited by training length. Both analyses involved batches of 256 samples. Model weights were updated by Adam optimizer \cite{kingma2014adam} using mean squared error as the metric to optimize. We employed early stopping during optimization; we stopped training if accuracy had not improved for 200 subsequent epochs. From each replicate, we extracted the minimal validation error and the corresponding model weights. These were the outputs considered during subsequent analyses. For $k$ dimensions that were equal to or greater than the underlying complexity, validation error was expected to not decrease further with $k$. To distinguish the ideal $k$ from overparameterized solutions, we chose the smallest $k$ for which the error was not statistically significant from the minimum error across all $k$ values.

\subsection{Global sensitivity analysis}
Following the acquisition of a latent representation, we determined the influence of the original parameters on the latent parameters using global sensitivity analysis.  We used the encoder (all layers before the bottleneck of the optimized neural network) to compute data pairs of input parameters and corresponding latent parameters.  Since these pairs were unevenly sampled due to our filtering steps, we employed structural and correlative sensitivity analysis (SCSA), a method that handles non-uniform sampling and accounts for correlations among inputs, to compute global sensitivities \cite{li2010global}. SCSA partitions sensitivity into uncorrelated and correlated contributions.  Of these, we used the uncorrelated component $S^{unc}_{ij}$, reflecting the exclusive contribution of each input parameter $x_i$, to more sparsely identify drivers of each latent variable $y_j$:

\begin{equation}
    S_{i,j}^{unc} = \sum_{s=1}^{N} (f_{p_{i,j}}(x_{i}^{(s)}))^2 / \sum_{s=1}^{N}(y_j^{(s)} - \bar{y_j})^2
\end{equation}

Where $s$ is the sample index, $N$ is the total number of samples, and $f_{p_{i,j}}$ is a data-driven sensitivity function for $x_i$ with respect to $y_j$ \cite{li2010global}. We used SALib's (v1.4.5) implementation of SCSA \cite{Herman2017, Iwanaga2022} to determine sensitivities. The above procedure was applied to every $j$-th latent parameter separately to derive each row of the $I \times J$ sensitivity matrix for $I$ input and $J$ latent parameters.

\subsection{Global fitting}
We employed global optimization to infer native and latent parameters from output data. In both cases, parameters were first normalized to a hypercube ($p_i \in [0, 1]$). For latent parameters, we determined bounds for the hypercube based on ranges that we observed for the latent parameters. Next, we used a basin-hopping algorithm (SciPy v1.7.1 \cite{virtanen2020scipy}) to find the best-fitting parameters. Basin-hopping combines a global step-taking routine and a local optimizer to find a global minimum across the parameter space. The global step-taking routine was initialized at 0.5 within the hypercube, and each step involved a random displacement of coordinates with a step size of 0.2. Local minima were found at each step, including the initial point, using the Broyden–Fletcher–Goldfarb–Shanno (BFGS) algorithm. The goodness of fit was evaluated based on the residual sum of squares. Steps were accepted with a probability $P$ determined by the change in the value of the cost function $f(p)$.

\begin{equation}
    P = exp(- \frac{f(x_{new}) - f(x_{old})}{T})
\end{equation}

Notice that the acceptance rate was 100\% for steps that improved the objective but still nonzero for steps that yielded worse objectives. This acceptance rate is tuned through a temperature parameter $T$ individually adjusted for each model. This allows the algorithm to explore the landscape within the hypercube and thus greatly increases the likelihood of finding the global optimum.

	
\section{Results}    
\begin{figure*}[!t]
    \begin{center}
	    \includegraphics[width=\linewidth]{
	    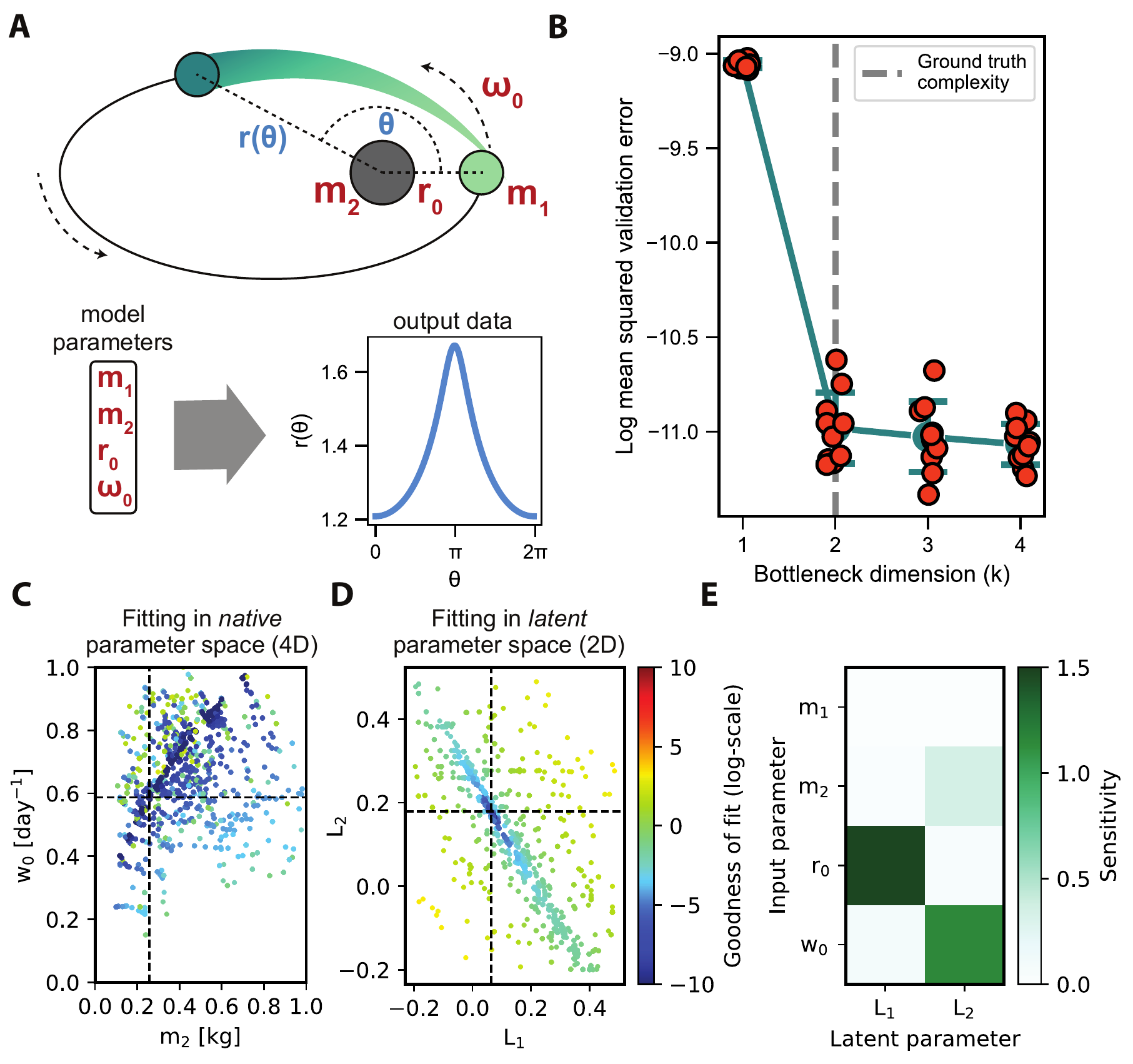}
    \end{center}
	\caption{\small\raggedright{{{\textbf{Recovering the known parameter redundancies of the Kepler orbit model.}
 \textbf{A}: Diagram of an example Keplerian orbit, the corresponding input parameters (red), and the model outputs ($\theta$, $r$ pairs).
 \textbf{B}: Validation error of the compressed representation found by FixFit as a function of the bottleneck dimension ($k$). Multiple replicates were performed at each $k$ using a stochastic optimizer. Shown for each $k$ are the individual data points, mean, and standard error. FixFit correctly identified the underlying redundancy of the Kepler orbit model by indicating saturating error at $k=2$ (see Methods). In subsequent panels, we utilized one of the fitted neural network replicates at $k=2$.
 \textbf{C}: Objective landscape of two of the original four dimensions. Each point corresponds to an optimizer evaluation and the corresponding objective value (log-transformed sum of squares error) with respect to two parameters, $m_2$ and $\omega_0$ (scaled between $0$ and $1$). The broad distribution of low error evaluations suggests a parameter redundancy. Consequently, standard fitting could not reliably identify the underlying ground truth parameters (marked by dashed lines).
 \textbf{D}: Objective landscape of the two latent dimensions. In contrast to the previous case, the same optimizer procedure converged to the correct, unique minimum in the latent parameter space ($L_1$,$L_2$) identified by FixFit. \textbf{E}: SCSA global sensitivities of the latent parameters to the four original parameters. Higher values of sensitivity (green) indicate a stronger influence. Considering the closed-form solution as a reference, SCSA correctly identified that parameter $m_1$ had no influence on outputs and that the remaining three parameters $m_2$, $r_0$, and $\omega_0$ together determined the two latent parameters.}}}}
	\label{fig:kepler}
\end{figure*}

\subsection{Recovering the known parameter redundancies of the Kepler orbit model}
The Kepler model \cite{bate2020fundamentals} describes the elliptical orbit (pairs of angles ($\theta$) and radii ($r$), see  \autoref{fig:kepler}A) of two gravitationally-attracting bodies as a function of four input parameters ($m_1$, $m_2$, $r_0$, $\omega_0$) (Methods). Ellipses, however, can be entirely described by two composite shape parameters, eccentricity ($e$) and the semi-latus rectum ($l$). As the dependencies between the four input and two composite parameters are known analytically, our results can be compared against a ground truth.

After generating simulated data using the original model, we trained a neural network with a variable-width bottleneck layer ($k=1...4$, (Figure S7 \cite{SI})) to determine, from data, the underlying complexity of the Kepler model (see Methods). The validation error saturated at $k=2$, suggesting that the underlying minimal representation, consistent with the ground truth, is two-dimensional (\autoref{fig:kepler}B). In subsequent analyses, we thus utilized a trained neural network from one of the replicates at $k=2$ to encode a minimal representation of the four input parameters.

\begin{figure*}[!t]
    \begin{center}
	    \includegraphics[scale=0.65]
     {
	    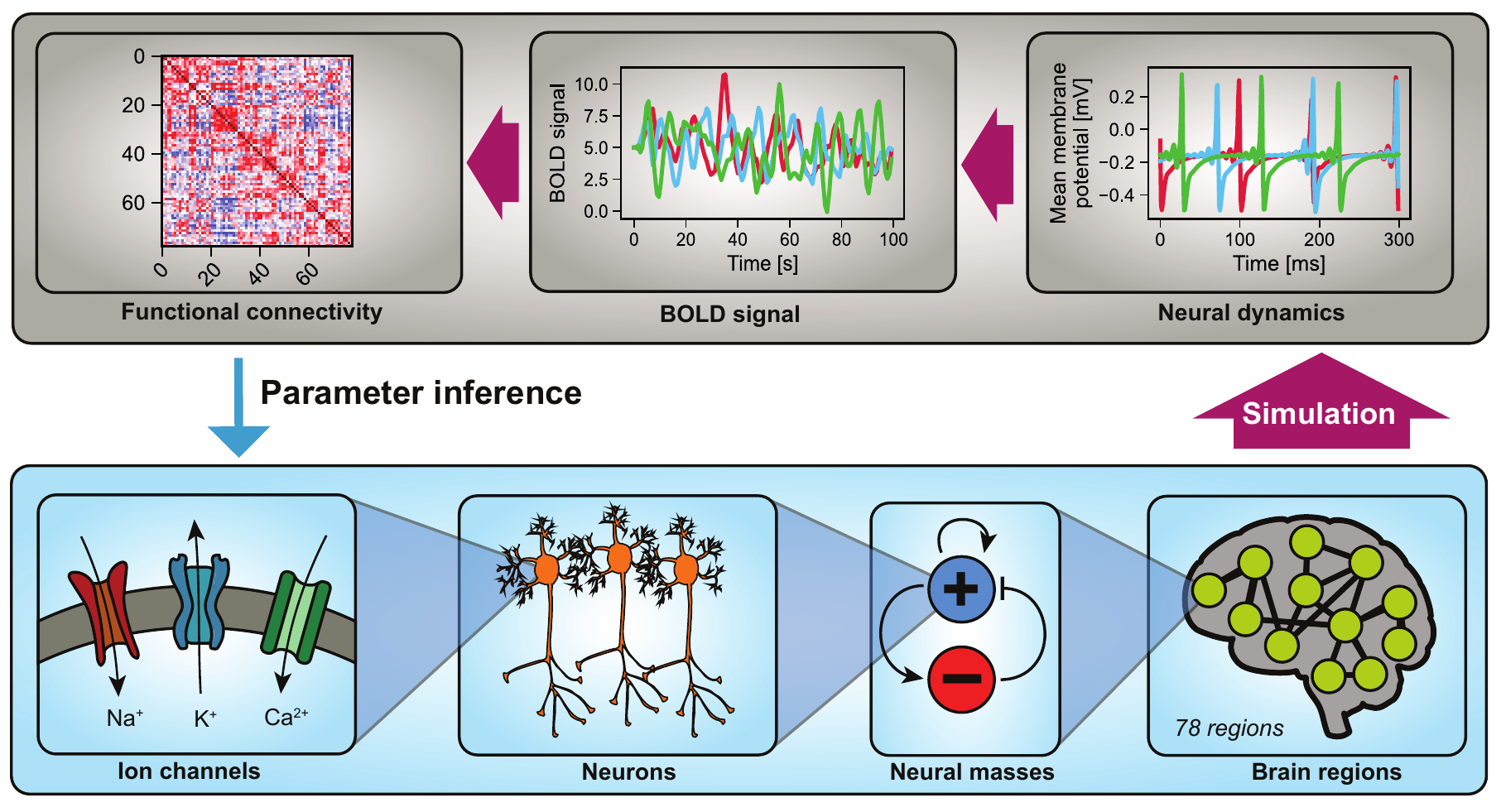}
    \end{center}
	\caption{\small\raggedright{{{\textbf{The Larter-Breakspear brain model connects microscopic properties of neurons with patterns of brain-wide activity.} This model produces brain dynamics by considering neuronal properties on multiple spatial scales (bottom). On the bottom scale, it takes into account transmembrane ion transport (bottom left), which is the basis of neuronal dynamics. Ion transport is facilitated by voltage-gated ion channels that are separately specified for $\text{K}^+$, $\text{Na}^+$, and $\text{Ca}^{2+}$ ions. Next, a mean-field approximation aggregates single-cell behavior into the population level of neural masses. An interacting pair of an excitatory (+) and an inhibitory (-) population is integratively modeled to form individual brain regions. Finally, on the whole-brain scale, a network of 78 connected brain regions (connected based on diffusion tensor imaging) is considered to complete the spatial span of the model (bottom right). The simulated region-specific brain dynamics are then transformed to achieve compatibility with functional MRI, a non-invasive neuroimaging modality (top). Simulated time series (top right) are first converted into blood-oxygen-level dependent (BOLD) signals using a hemodynamic response function. Then, to remove phase-specific information, all-to-all Pearson correlations are computed among the time series resulting in a 78-by-78 functional connectivity (FC) matrix (top left).}}}}
	\label{fig:lbscheme}
\end{figure*}

To demonstrate the practical issue of inferring degenerate model parameters, we attempted to uniquely infer all four original parameters from simulated Kepler orbits. As expected, the global optimizer found a wide set of equally optimal solutions across the parameter space (\autoref{fig:kepler}C). By contrast, global optimization in latent parameter space (Figure S8 \cite{SI}) (through the decoder section of the neural network, see Methods) recovered a unique solution that coincided with the expected ground truth (\autoref{fig:kepler}D).

Next, we demonstrate that, by employing Structural and Correlative Sensitivity Analysis (SCSA), FixFit can quantify relationships between the original and latent model parameters (see Methods). Applied to the Kepler model (\autoref{fig:kepler}E), the approach provides global sensitivities consistent with closed-form solutions for the two intermediate terms $e$ and $l$. Firstly, as expected, no sensitivity was detected with respect to $m_1$. As such, $m_1$ was a completely redundant parameter. The remaining three parameters, $r_0$, $m_2$ and $\omega_0$ together determined $L_1$. Finally, $r_0$ alone influenced $L_2$. Both ground truth terms, $e$ and $l$, by contrast, included all three of these parameters (equations \ref{eq:kepler_e_short} and \ref{eq:kepler_l_short}). However, these analytical expressions share a common term with respect to the three input parameters and differ only by an additional power of $r_0$ in $l$.  By exploiting this relationship, FixFit separates the influence of $r_0$ from that of $m_2$ and $\omega_0$, thus providing a sparser latent representation compared to the ground truth.

\subsection{Identifying novel parameter relationships in a multi-scale brain model}
The Larter-Breakspear model is commonly used to connect microscopic neuronal properties with emergent brain activity, such as that measured using functional magnetic resonance imaging (fMRI) \cite{logothetis2008we} (\autoref{fig:lbscheme}). This is achieved by modeling the voltage-gated ion dynamics (here, $\text{Na}^{+}$, $\text{K}^{+}$, and $\text{Ca}^{2+}$) of a population of neurons called a ``neural mass'' \cite{larter1999coupled, breakspear2003modulation, chesebro2023ion}. Previous studies have further shown that, by coupling multiple neural masses according to the inter-regional connectivities measured through Diffusion Tensor Imaging (DTI), one can simulate semi-realistic fMRI dynamics \cite{endo2020evaluation}. As a challenge, this model has many parameters such that even after assigning identical parameter values to all 78 of the resulting brain regions and fixing biologically inert parameters, the model still has eleven remaining parameters corresponding to coupled biological processes (see Methods). Furthermore, these simulated fMRI activities are typically studied using functional connectivity (FC) or the matrix of region-to-region Pearson correlations \cite{biswal1995functional}, thus resulting in a reduction of information. To address this issue, we applied FixFit to resolve potential redundancies and, more broadly, characterize new relationships among the parameters. Microscopic model parameters, such as ion channel conductivity,  are often implicated in disease mechanisms and potential therapeutic targets, which makes their determination valuable \cite{de2014brain}.

\begin{figure*}[!t]
    \begin{center}
	    \includegraphics[scale=0.65]
     {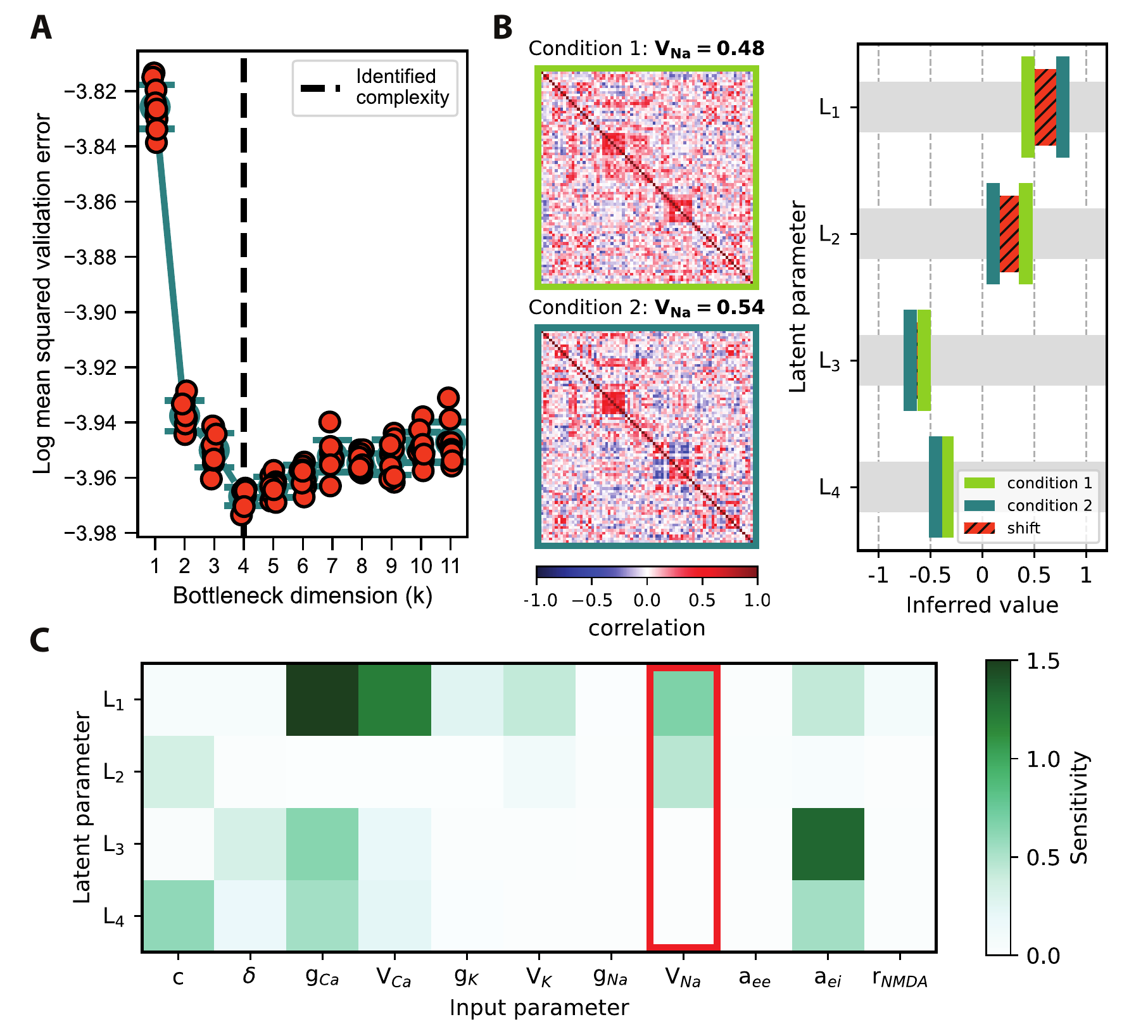}
    \end{center}
	\caption{\small\raggedright{{{\textbf{
Identifying novel parameter relationships in a multi-scale brain model.}
 \textbf{A}: Validation error of the Larter-Breakspear model as a function of the bottleneck dimension $(k)$. Shown for each $k$ are the individual data points, mean, and standard error. The minimal error occurs at $k=4$, implying that the eleven model parameters can be summarized by four latent parameters while still uniquely mapping to output space.
 \textbf{B}: Simulated functional connectivity matrices (left) and corresponding fitting results in latent space (right) for two different values of $V_\text{Na}$.  Fitting converged to global minima in both conditions and indicated shifts in $L_1$ and $L_2$ (right, orange).
 \textbf{C}: Global sensitivities, as computed through SCSA, of the four latent parameters to the 11 original model parameters.  Latent parameter sensitivities to input parameters are color-coded. Three input parameters, $g_\text{Na}$, $a_{ee}$, and $r_{\text{NMDA}}$, do not appear to influence model outputs. $V_\text{Na}$ (marked with a red bracket), which we investigated in our perturbation experiment, influenced latent parameters $L_1$ and $L_2$ but not $L_3$ and $L_4$. These sensitivities are consistent with our fitting results in panel B, where we only detected shifts in $L_1$ and $L_2$.}}}}
	\label{fig:lbresults}
\end{figure*}

As with the previous example, we first trained neural networks with various bottleneck widths (Methods) to quantify the information content of our fMRI simulations (Figure S9 \cite{SI}). The minimum validation error occurred at $k=4$ (\autoref{fig:lbresults}A), suggesting that four latent parameters ($L_1$, $L_2$, $L_3$, $L_4$) capture the composite effect of the original eleven parameters.

Using a representation acquired at $k=4$, we next give an example of how to infer latent parameters from model outputs and interpret the results in input space. For this purpose, we first perturbed the $\text{Na}^+$ reversal potential ($V_\text{Na}$) and considered two conditions ($V_\text{Na}=\{0.48,0.54\})$ while keeping all other parameters fixed (\autoref{fig:lbresults}B). We observed that increasing $V_\text{Na}$, on average, reduced all correlations and introduced more anti-correlations across the brain. We then performed parameter inference in latent space in both conditions (Figure S10 \cite{SI}), and in each case, the global optimizer converged to a single unique solution. The two solutions differed only along the $L_1$ and $L_2$ axes, whereas detected changes with respect to $L_3$ and $L_4$ were minimal.

To interpret the observed shifts in the fitted latent parameters, we again applied global sensitivity analysis (\autoref{fig:lbresults}C). We found three parameters, $g_\text{Na}$, $a_{ee}$, and $r_{\text{NMDA}}$ had no effect on the latent parameters and therefore did not affect FC model outputs. Since no shifts in this example were observed in $L_3$ and $L_4$, 5/8 of the remaining parameters can be eliminated. This reduced the candidate parameters from eleven down to three, with $V_\text{Na}$ contributing to the shift in $L_1$ and $L_2$ (\autoref{fig:lbresults}C) (consistent with \autoref{fig:lbresults}B). As demonstrated by this example, FixFit can be used to narrow down potential mechanisms for the observed changes in latent parameters. This reduction makes the remaining uncertainty tractable to resolve in specialized experiments that target individual parameters on the single neuron scale.

\section{Discussion}
While many methods exist for fitting model parameters to data, most of them are limited to situations where there is a need for a single, optimal solution. However, many different parameter sets often fit the data equally well due to complex parameter dependencies and the limited information content of experimental measurements \cite{cole2020parameter}. Therefore, to fully assess what model features data can and cannot resolve, we need tools that describe the interaction between these solutions. In this paper, we have presented FixFit, a data-driven approach that uniquely identifies a minimal set of latent features from the set of interacting model parameters. Our approach correctly identified that, as previously established, two latent parameters were sufficient to characterize the Kepler orbit model. Furthermore, we demonstrated in the Larter-Breakspear brain network model that we can use changes in such composite features to narrow down parameter candidates that shift in response to experimental perturbations.

To validate FixFit, we used two synthetic use cases; however, our framework can be adapted to infer parameters from new experimental data. Crucially, experimental data can include a significant amount of measurement noise which can introduce additional redundancies among model parameters. Our approach can be used to gauge how resolvable complexity depends on the amount of noise. Various methods are available to supplement simulations with noise and yield realistic model outputs \cite{scales1998noise, fuller2009introduction, gravel2004method, goldwyn2011and}. 

The bottleneck layer within FixFit finds a minimal and sufficient latent representation without any additional constraints \cite{achille2017emergence}, but, as with other dimensionality reduction methods, \cite{kaiser1958varimax, xu2003document}, the representation is not unique. This, however, provides FixFit with an additional layer of flexibility that can be exploited. For example, variance maximization (Varimax) or sparsity constraints may be placed during training to select, among equally accurate alternatives, representations where latent variables are functions of the fewest possible input parameters \cite{kaiser1958varimax, ghasedi2017deep}.

The idea of achieving a minimal and sufficient representation of a physical system by implementing compression on dynamical systems has been previously explored \cite{chen2022automated}. Our approach builds upon this notion by linking the compressed representation to model parameters, which enables us to address redundancies in the original parameter space. However, to ensure the utility of this representation discovered by our tool, it must be interpreted in relation to the input parameters. This can be a challenge due to the complexity of the encoder that maps between these two elements. Various methods are available to interpret the components of an already-trained neural network. These include sensitivity analysis \cite{zhang2019novel}, feature importance scoring \cite{murdoch2019definitions}, activation maximization \cite{montavon2018methods}, and gradient-based saliency mapping \cite{selvaraju2017grad}. Thanks to the modular structure of our approach, it has the advantage of adaptability and can be interpreted using any one of these methods.

For the demonstration of our framework, we used SCSA, a sensitivity analysis variant, to quantify the impact of each model parameter on the latent parameters \cite{li2010global}. However, as with the other mentioned approaches, sensitivity analysis does not provide explicit formulas for the nonlinear parameter relationships identified by the neural network. Besides revealing more information about relationships between input parameters, such formulas would allow latent parameters to be understood as composite parameters that may be easier to interpret (i.e., eccentricity in the Kepler orbit model). This, in turn, could allow FixFit to develop intuitive, simplified models. Symbolic regression offers a possible avenue to achieve this by approximating the encoder \cite{schmidt2009distilling, la2021contemporary} or by itself being incorporated into the neural network \cite{petersen2019deep, kim2020integration}.

Bayesian techniques are also a viable means of inferring parameters, with the benefit of providing uncertainties for the estimates, even in the presence of moderate degeneracies \cite{carlin1995bayesian, jaakkola2000bayesian, gonccalves2020training}. Nevertheless, they are prone to fail in cases of pronounced structural non-identifiability, where due to bad mixing of the sampling algorithm, the true values are not captured \cite{wieland2021structural}. In addition, assessing interplay among parameters from posteriors becomes challenging in high dimensional cases. In contrast, our approach adeptly handles structural non-identifiability as the latent representation ensures interpretability even for high dimensional models, such as the eleven-parameter brain network model we show here. Although distinct in their approaches, Bayesian techniques and our approach could be combined synergistically to harness their individual strengths. Once the low dimensional latent representation is established, Bayesian methods could be applied efficiently to consider priors and quantify uncertainties of the latent parameters in the presence of noise.

Future studies could utilize our framework to rank different experimental designs based on how well they resolve specific parameters of interest in simulations.  In neuroscience, for example, brain-wide activities can be measured in a multitude of ways, including by fMRI, electroencephalography (EEG), and magnetoencephalography (MEG). FixFit could use simulations of these modalities from models (e.g., Larter-Breakspear) to determine which parameters each can resolve and, as a result, when to use them. Further, our framework can suggest targeted follow-up experiments to augment pre-existing data and inferred latent parameters. Related methods of optimal design aim to select experimental conditions such as temperatures, concentrations, and sample sizes to most easily resolve the parameters of a fixed model \cite{chaloner1995bayesian, liepe2013maximizing, smucker2018optimal}. Thus, our tool could work synergistically with these prior approaches to optimize future data collection.

\section{Conclusion}
In conclusion, we present FixFit, a method to identify unique representations of redundant model parameters of computational models. Our tool enables scientists to select the most informative measurement modalities for their experiments, particularly in fields such as neuroscience, where it is still ambiguous what information experiments can resolve. Additionally, using our approach, researchers can utilize an extensive library of existing computational models for parameter inferences. These efforts can be further enhanced by specialized experiments that address the remaining parameter uncertainties, leading to greater synergy between modeling and experimentation.

\section*{Code availability}
All relevant code and data can be accessed at www.lcneuro.org/analytic/fixfit.

\section*{Acknowledgements}
Research presented here was funded by the following:  National Science Foundation/White House Brain Research Through Advancing Innovative Technologies (BRAIN) Initiative, United States (NSFNCS-FR 1926781, LRMP), Baszucki Brain Research Fund, United States (LRMP), NIHGM MSTP Training Award, United States (T32-GM008444, AGC), and the Marie-Josée Kravis Fellowship in Quantitative Biology (CW). In addition, we would like to thank David Hofmann for valuable discussions.


\bibliography{main}

\end{document}


\title{FixFit: deep learning for optimal model reduction}

\author{
  \IEEEauthorblockN{Botond B. Antal\IEEEauthorrefmark{1}\IEEEauthorrefmark{2}},
  \IEEEauthorblockN{Anthony G. Chesebro\IEEEauthorrefmark{1}\IEEEauthorrefmark{3}},
  \IEEEauthorblockN{Helmut H. Strey\IEEEauthorrefmark{1}\IEEEauthorrefmark{4}},
  \IEEEauthorblockN{Lilianne R. Mujica-Parodi\IEEEauthorrefmark{1}\IEEEauthorrefmark{2}\IEEEauthorrefmark{3}\IEEEauthorrefmark{4}},
  \IEEEauthorblockN{Corey Weistuch\IEEEauthorrefmark{5}}\\
  
  \IEEEauthorblockA{\IEEEauthorrefmark{1}Department of Biomedical Engineering, Stony Brook University, Stony Brook, NY, USA}\\
  \IEEEauthorblockA{\IEEEauthorrefmark{2}Athinoula A. Martinos Center for Biomedical Imaging, Massachusetts General Hospital and Harvard Medical School, Boston, MA, USA}\\
  \IEEEauthorblockA{\IEEEauthorrefmark{3}Renaissance School of Medicine, Stony Brook University, Stony Brook, NY, USA}\\
  \IEEEauthorblockA{\IEEEauthorrefmark{4}Laufer Center for Physical and Quantitative Biology, Stony Brook University, Stony Brook, NY, USA}\\
  \IEEEauthorblockA{\IEEEauthorrefmark{5}Department of Medical Physics, Memorial Sloan Kettering Cancer Center, New York, NY, USA}\\

\thanks{Email: weistucc@mskcc.org}
}

\date{\today}
\maketitle

\justifying
\section{Supplementary information}
\begin{figure}[htb]
    \begin{center}
	    \includegraphics[scale=1]{
	    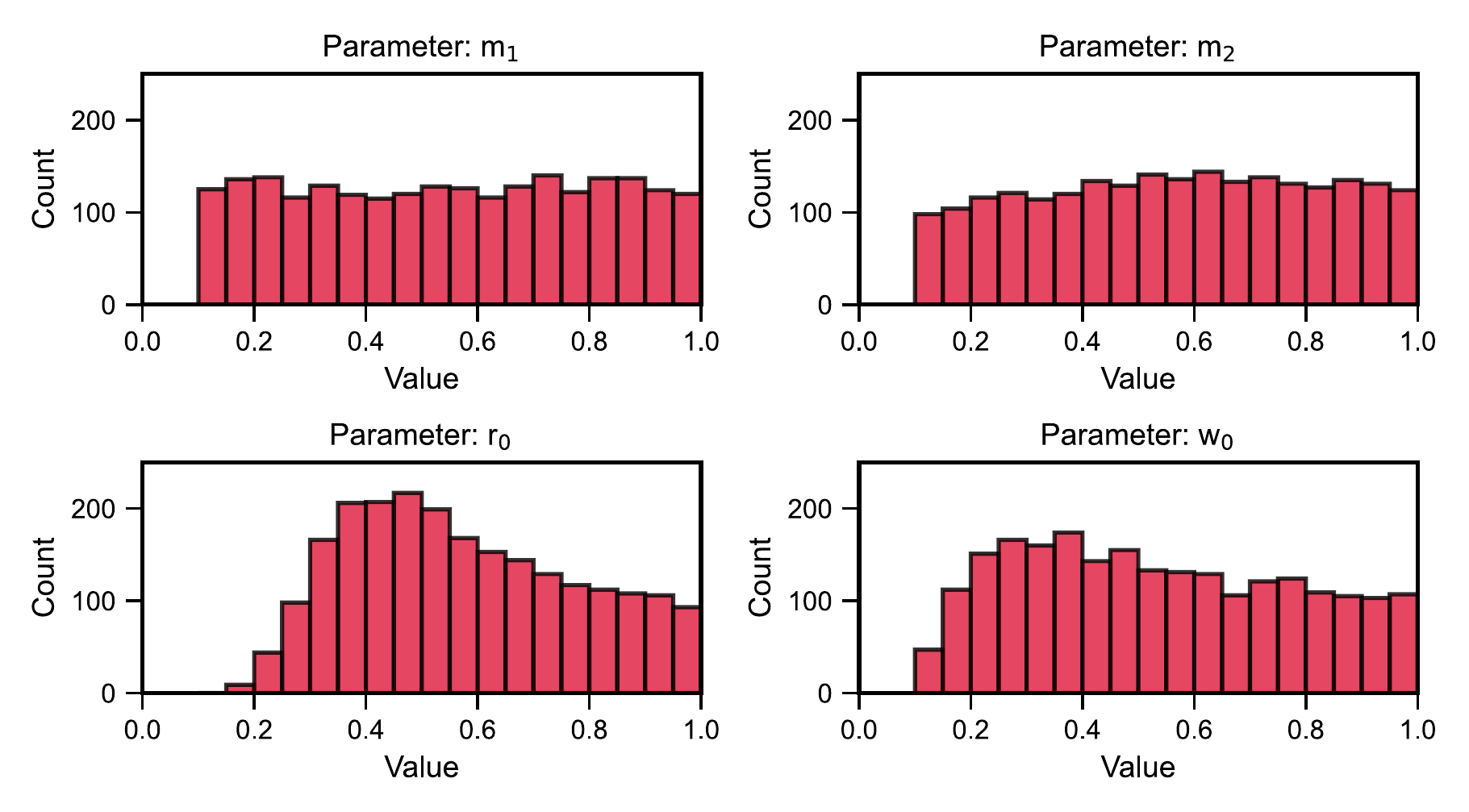}
    \end{center}
	\caption*{Fig. S1:
 \textbf{Kepler orbit model: distributions of input parameters in the training dataset.}
 The Kepler model had four input parameters. These were sampled for the training dataset using a Sobol sequence algorithm in four dimensions. The sampled parameter sets were then filtered based on eccentricity criteria. The above histograms show the resultant distributions for the four input parameters from 2,276 samples that passed these criteria.
 }
	\label{fig:kep_inp}
\end{figure}

\clearpage
\begin{figure}
    \begin{center}
	    \includegraphics[scale=1]{
	    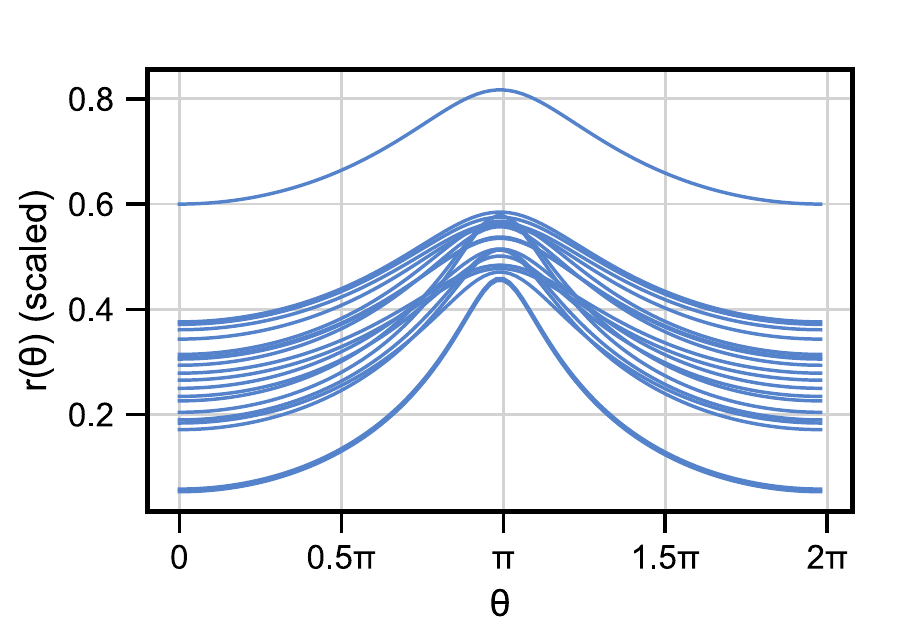}
    \end{center}
	\caption*{Fig. S2:
 \textbf{Kepler orbit model: examples of model outputs from the training dataset.}
 Output data for the Kepler model entailed polar coordinate representation of planetary orbits. For each sample, we first computed $r(\theta)$ at 100 $\theta$ increments, given the input parameters, and then we log-transformed the computed $r(\theta)$ values. Finally, the full set of outputs was scaled to [0, 1]. The above-shown curves are 20 randomly chosen examples of output data from the training dataset.
 }
	\label{fig:kep_out}
\end{figure}

\clearpage
\begin{figure}[htb]
    \begin{center}
	    \includegraphics[scale=1]{
	    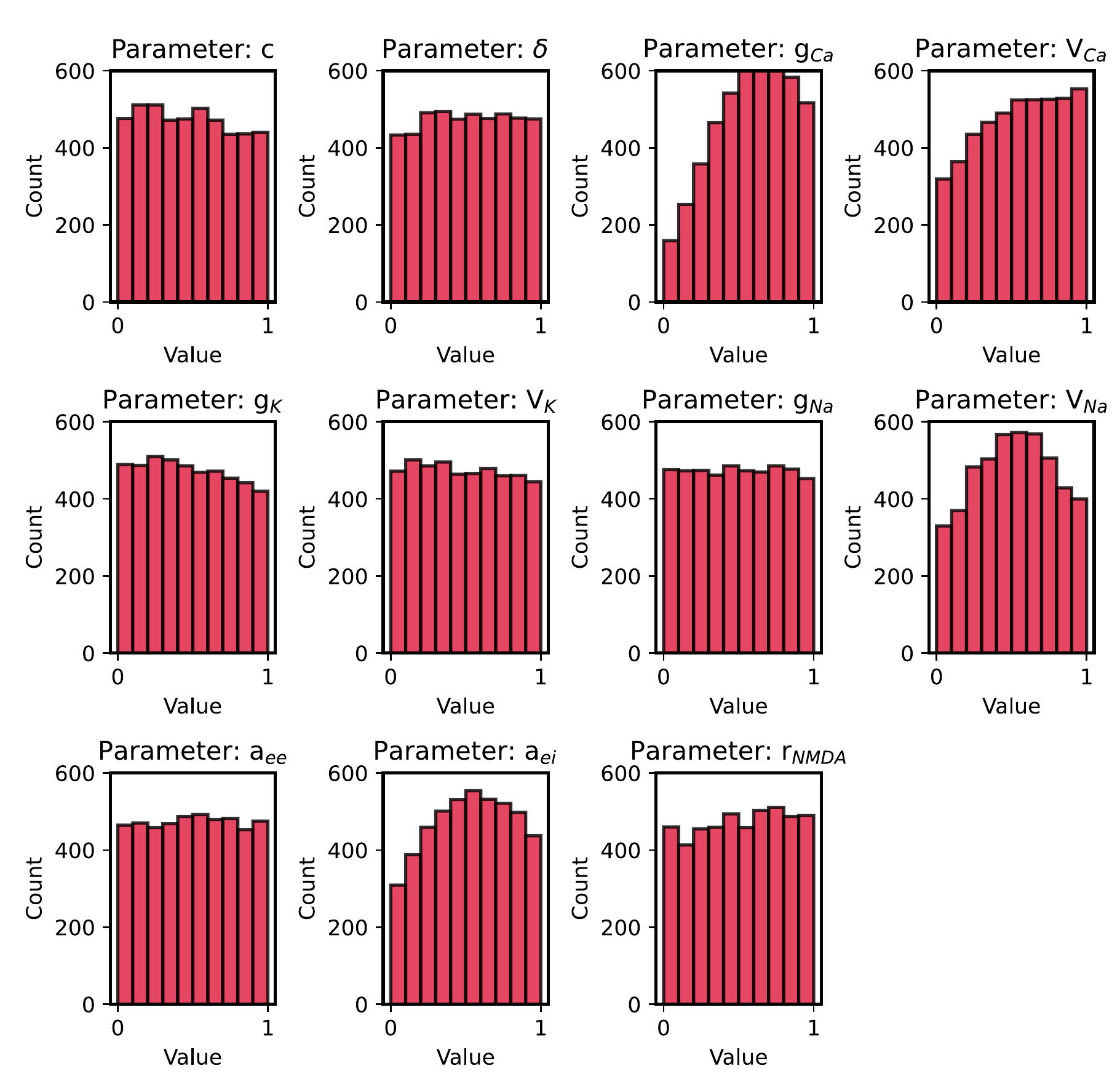}
    \end{center}
	\caption*{Fig. S3:
 \textbf{Brain network model: distributions of input parameters in the training dataset.}
 The Larter-Breakspear brain network model had eleven input parameters. These were sampled for the training dataset using a Sobol sequence algorithm in eleven dimensions given predefined ranges. The sampled parameter sets were then filtered based on criteria that tested for oscillatory behavior and constrained the average functional connectivity. Finally, the input parameters were scaled to [0, 1] before training. The above histograms show the resultant distributions for the eleven input parameters from 4,730 samples that passed the described criteria.
 }
	\label{fig:lb_inp}
\end{figure}

\clearpage
\begin{figure}[htb]
    \begin{center}
	    \includegraphics[scale=1]{
	    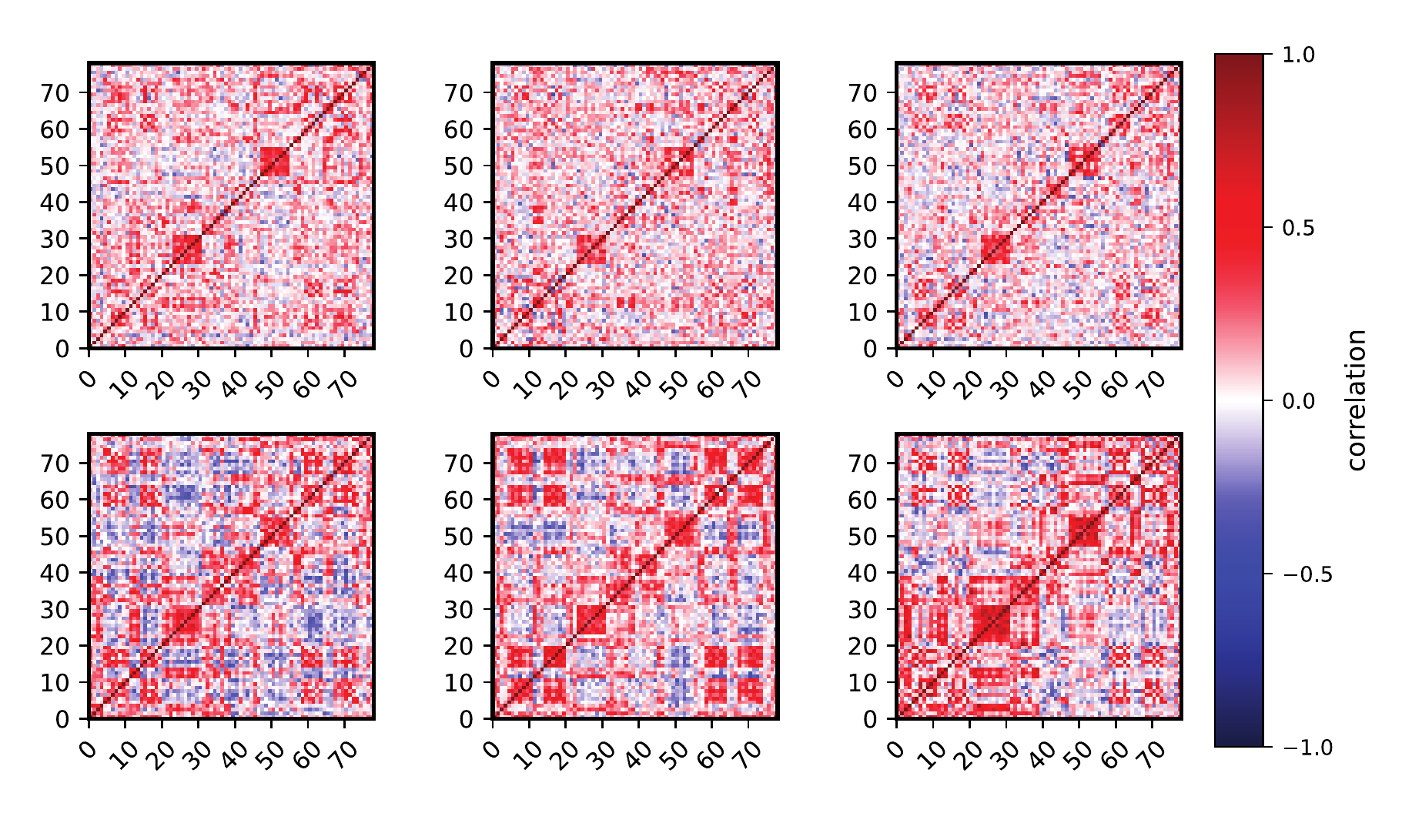}
    \end{center}
	\caption*{Fig. S4:
 \textbf{Brain network model: examples of model outputs from the training dataset.}
 Six examples of functional connectivity matrices from the training dataset are shown above for the Larter-Breakspear brain network model. Functional connectivity was quantified among 78 brain regions using Pearson correlations. Since the resultant values were between -1 and 1, no additional scaling was implemented for the neural network training.}
	\label{fig:lb_out}
\end{figure}

\clearpage
\begin{figure}[htb]
    \begin{center}
	    \includegraphics[scale=1]{
	    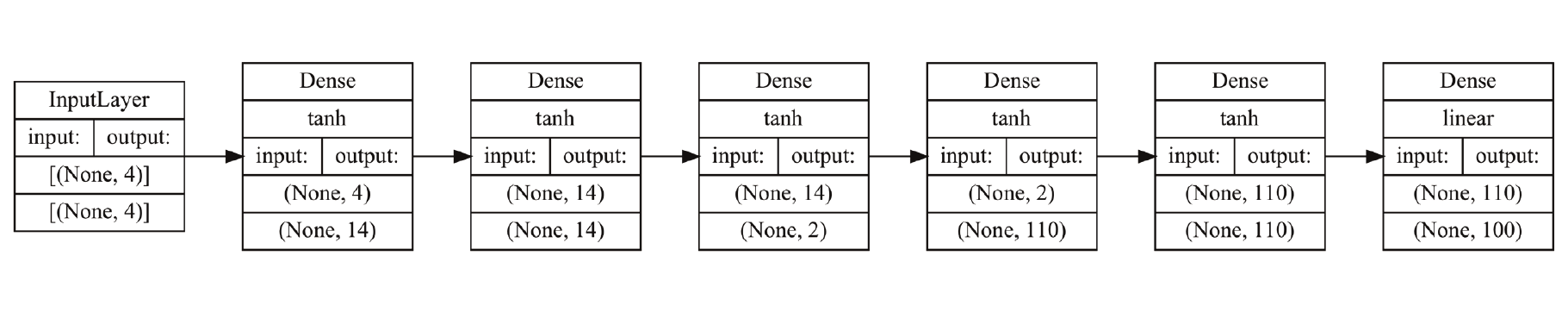}
    \end{center}
	\caption*{Fig. S5:
 \textbf{Kepler orbit model: neural network architecture.}
 The above diagram describes the architecture of the neural network we applied to the Kepler orbit model. The neural network involved fully connected layers, among which one of the intermediate layers was a bottleneck layer. We retrained this architecture while varying the dimensionality of the bottleneck layer to identify the underlying complexity of the model.
 }
	\label{fig:kep_nn}
\end{figure}

\clearpage
\begin{figure}[htb]
    \begin{center}
	    \includegraphics[scale=1]{
	    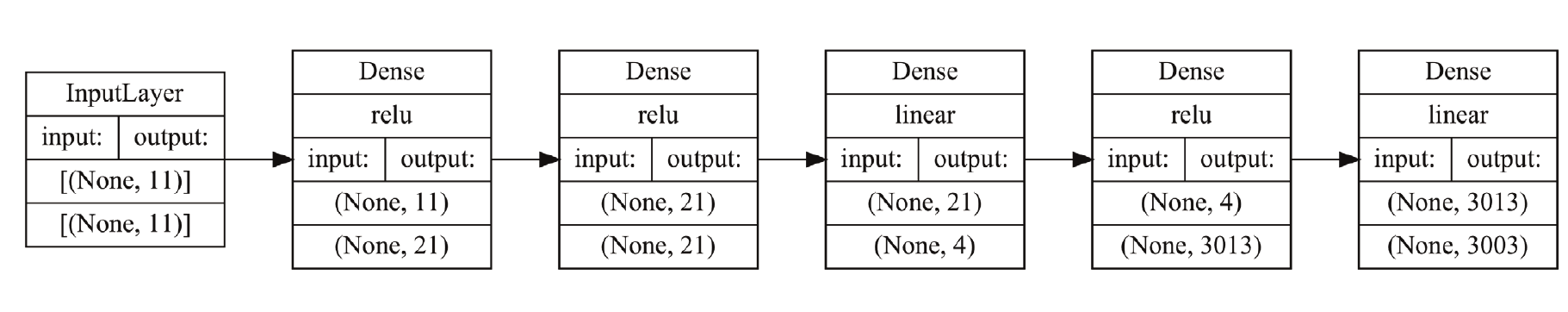}
    \end{center}
	\caption*{Fig. S6:
 \textbf{Brain network model: neural network architecture.}
 The above diagram describes the architecture of the neural network we applied to the brain network model. The neural network involved fully connected layers, among which one of the intermediate layers was a bottleneck layer. We retrained this architecture while varying the dimensionality of the bottleneck layer to identify the underlying complexity of the model.
 }
	\label{fig:lb_nn}
\end{figure}

\clearpage
\begin{figure}[htb]
    \begin{center}
	    \includegraphics[scale=1]{
	    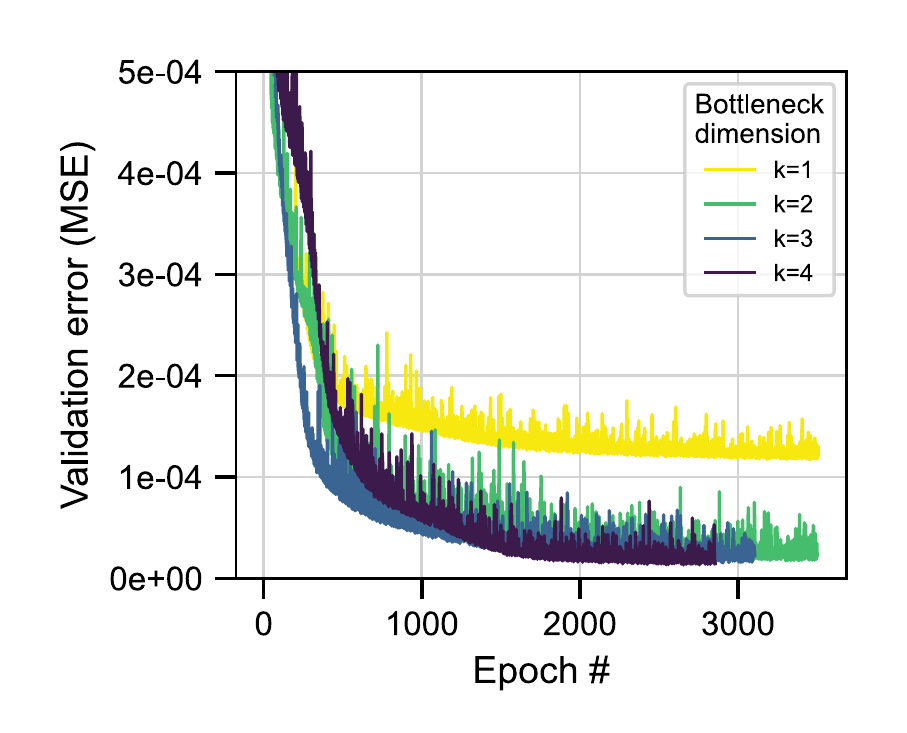}
    \end{center}
	\caption*{Fig. S7:
 \textbf{Kepler orbit model: evolution of error during neural network training.}
 The above curves show how error on a validation dataset progressed throughout the training of the applied neural network. The error was quantified as mean-squared error and is displayed for a single replicate from each tested bottleneck dimension. Early stopping was implemented to avoid overfitted models: the training was stopped if there had not been an improvement for the 200 most recent epochs. The final model was selected from the epoch with minimal error.
 }
	\label{fig:kep_train}
\end{figure}

\clearpage
\begin{figure}[htb]
    \begin{center}
	    \includegraphics[scale=1]{
	    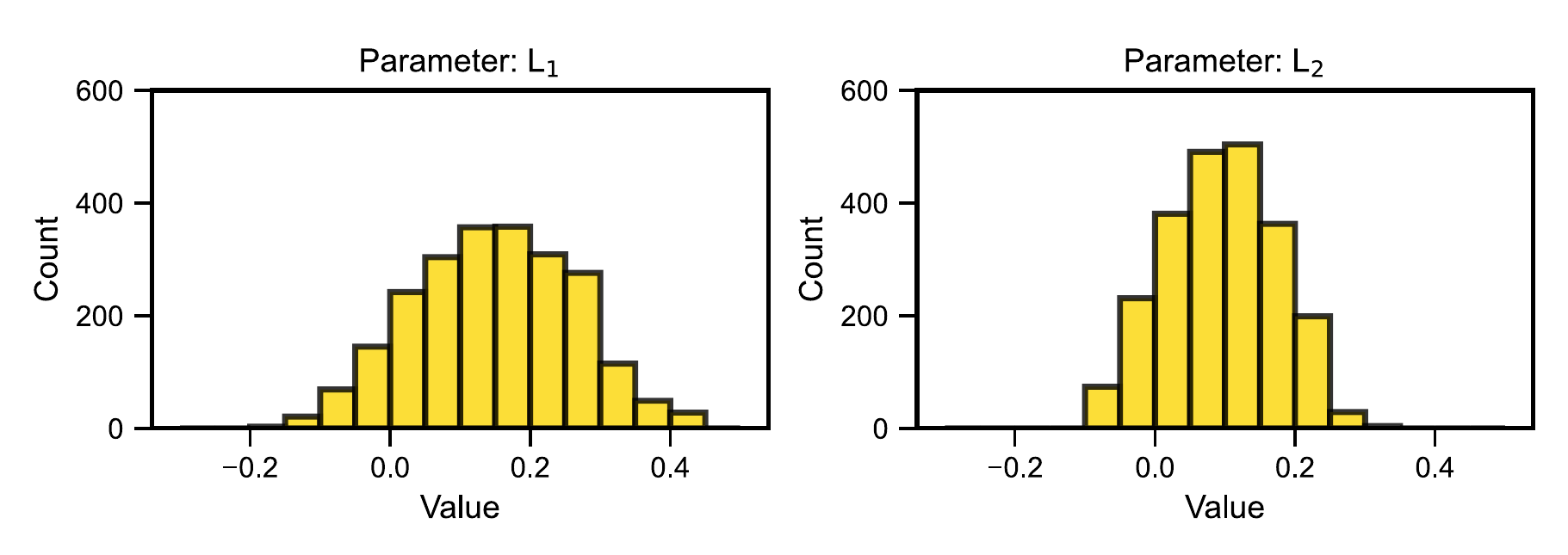}
    \end{center}
	\caption*{Fig. S8:
 \textbf{Kepler orbit model: distribution of latent parameters following training.}
 After we had trained the neural networks and selected the minimal dimensionality for the bottleneck, we computed the values of the latent representation for the training dataset. We acquired the values of the two latent parameters from the bottleneck layer, and we display their distributions above. These distributions were considered when choosing bounds for latent parameters for the global fitting.
 }
	\label{fig:kep_latent}
\end{figure}

\clearpage
\begin{figure}[htb]
    \begin{center}
	    \includegraphics[scale=1]{
	    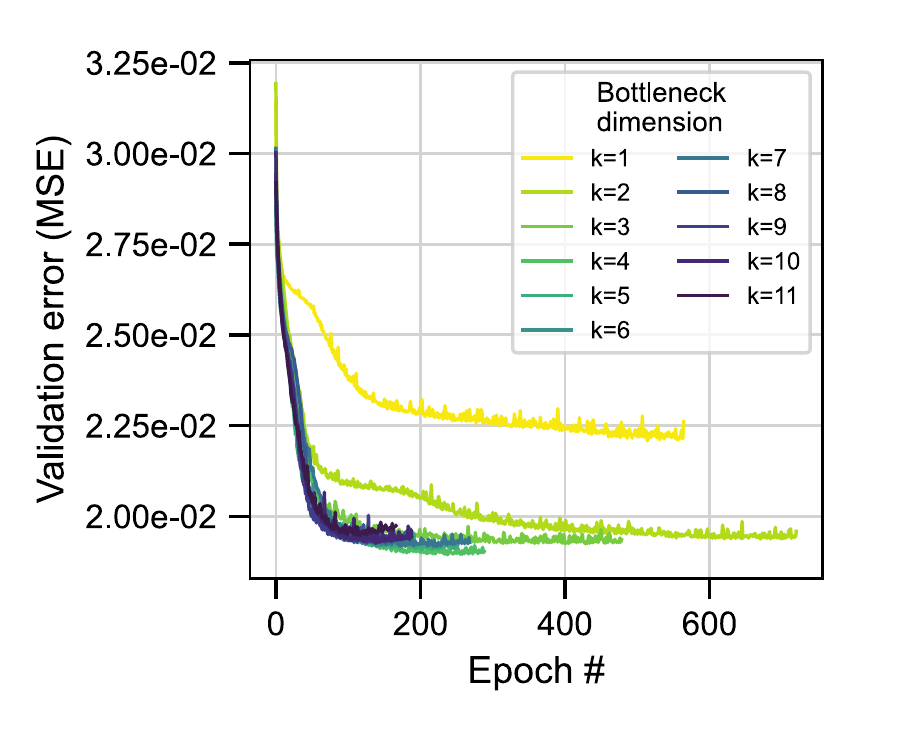}
    \end{center}
	\caption*{Fig. S9:
 \textbf{Brain network model:  evolution of error during neural network training.}
 The above curves show how error on a validation dataset progressed throughout the training of the applied neural network. The error was quantified using mean-squared error and is displayed for a single replicate from each tested bottleneck dimension. Early stopping was implemented to avoid overfitted models: the training was stopped if there had not been an improvement for the 200 most recent epochs. The final model was selected from the epoch with minimal error.
 }
	\label{fig:lb_train}
\end{figure}

\clearpage
\begin{figure}[htb]
    \begin{center}
	    \includegraphics[scale=1]{
	    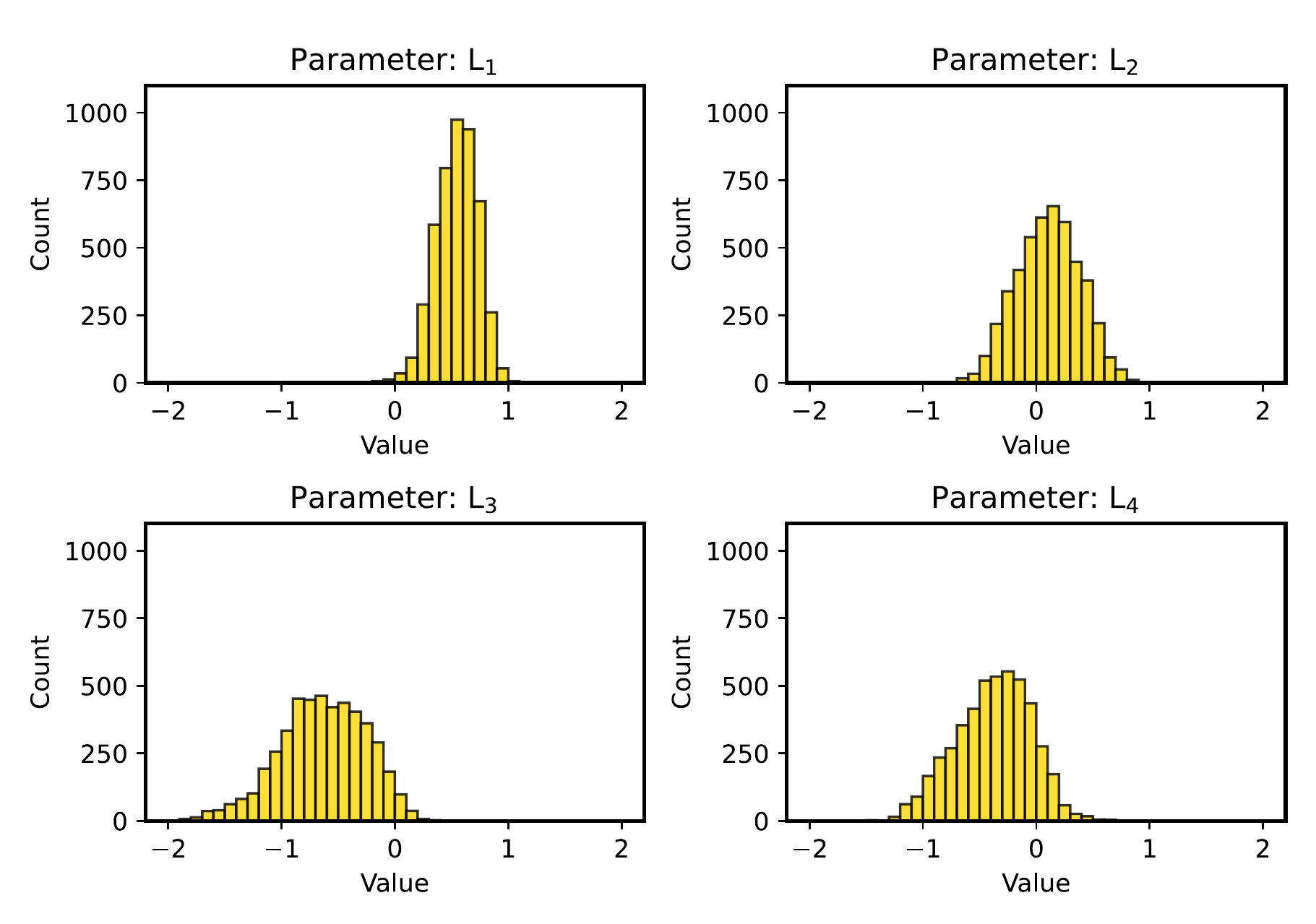}
    \end{center}
	\caption*{Fig. S10: \textbf{Brain network model:  distribution of latent parameters following training.}
  After we had trained the neural networks and selected the minimal dimensionality for the bottleneck, we computed the values of the latent representation for the training dataset. We acquired the values of the four latent parameters from the bottleneck layer, and we display their distributions above. These distributions were considered when choosing bounds for latent parameters for the global fitting.
 }
	\label{fig:lb_latent}
\end{figure}